\pdfoutput=1

\documentclass[11pt]{article}

\usepackage[final]{acl}
\usepackage{times}
\usepackage{latexsym}
\usepackage{booktabs}
\usepackage{amsmath}
\usepackage{color, colortbl}
\usepackage{tabularx}
\usepackage{paralist}
\definecolor{Gray}{gray}{0.9}
\usepackage[T1]{fontenc}
\usepackage[utf8]{inputenc}
\usepackage{microtype}
\usepackage{inconsolata}
\usepackage{graphicx}
\usepackage{multirow}
\usepackage{subcaption}
\usepackage{listings}
\usepackage{xcolor}

\usepackage{siunitx}

\lstdefinelanguage{XML}{
  morestring=[b]",
  moredelim=[s][\color{black}]{>}{<},
  moredelim=[s][\bfseries\color{blue}]{<}{\ },
  morecomment=[s]{<!--}{-->},
  commentstyle=\color{gray},
  stringstyle=\color{teal},
  identifierstyle=\color{black}
}

\title{Can LLMs Judge Debates?\\
Evaluating Non-Linear Reasoning via Argumentation Theory Semantics}

\author{
Reza Sanayei$^{1}$, 
Srdjan Vesic$^{2}$, 
Eduardo Blanco$^{1}$, 
Mihai Surdeanu$^{1}$ \\
$^{1}$Department of Computer Science, University of Arizona \\
$^{2}$CRIL CNRS \& University of Artois \\
\texttt{\{rsanayei, eduardoblanco, msurdeanu\}@arizona.edu}, 
\texttt{vesic@cril.fr}
}

\begin{document}
\maketitle
\begin{abstract}
Large Language Models (LLMs) excel at linear reasoning tasks but remain underexplored on non-linear structures such as those found in natural debates, which are best expressed as argument graphs. We evaluate whether LLMs can approximate structured reasoning from Computational Argumentation Theory~(CAT). Specifically, we use Quantitative Argumentation Debate (QuAD) semantics, which assigns acceptability scores to arguments based on their attack and support relations. Given only dialogue-formatted debates from two NoDE datasets, models are prompted to rank arguments without access to the underlying graph. We test several LLMs under advanced instruction strategies, including Chain-of-Thought and In-Context Learning. While models show moderate alignment with QuAD rankings, performance degrades with longer inputs or disrupted discourse flow. Advanced prompting helps mitigate these effects by reducing biases related to argument length and position. Our findings highlight both the promise and limitations of LLMs in modeling formal argumentation semantics and motivate future work on graph-aware reasoning.
\end{abstract}

\section{Introduction}

Evaluating the reasoning capabilities of Large Language Models (LLMs) has largely focused on tasks involving \textit{linear} reasoning, such as arithmetic, logical puzzles, and Chain-of-Thought (CoT) explanations~\cite{NEURIPS2022_8bb0d291}. Even complex inference methods—like beam search, best-of-\textit{N}, or Tree-of-Thoughts~\cite{yao2023tree}—are ultimately decomposable into linear sequences of intermediate steps. Similarly, research on conversational systems predominantly addresses linear interactions between users and LLMs.

In contrast, natural debates are not purely linear. 
Alongside serial (near-path) exchanges, they frequently exhibit \emph{branching} interactions, with multiple arguments converging on one claim (fan-in) and single arguments influencing several others (fan-out). We focus on these non-linear settings and use debates that exhibit substantial branching rather than simple chains. 
Computational Argumentation Theory (CAT) formalizes these complex interactions through frameworks like Quantitative Argumentation Debate (QuAD) semantics~\cite{baroni2018handbook}.
QuAD semantics assigns each argument a numerical \textit{acceptability degree} that reflects its strength within the debate.

The emerging paradigm of LLM-as-a-Judge~\cite{li2025generationjudgmentopportunitieschallenges, gu2025surveyllmasajudge} highlights the significant potential of LLMs for nuanced assessment and moderation across various applications, such as content evaluation, alignment, retrieval, and reasoning. Given the rapid advancement of multi-agent AI systems engaging in sophisticated argumentative interactions, it is increasingly crucial to understand whether LLMs can effectively serve as impartial judges capable of handling non-linear argumentation structures.

\begin{figure*}[t]
    \centering
    \includegraphics[width=\textwidth]{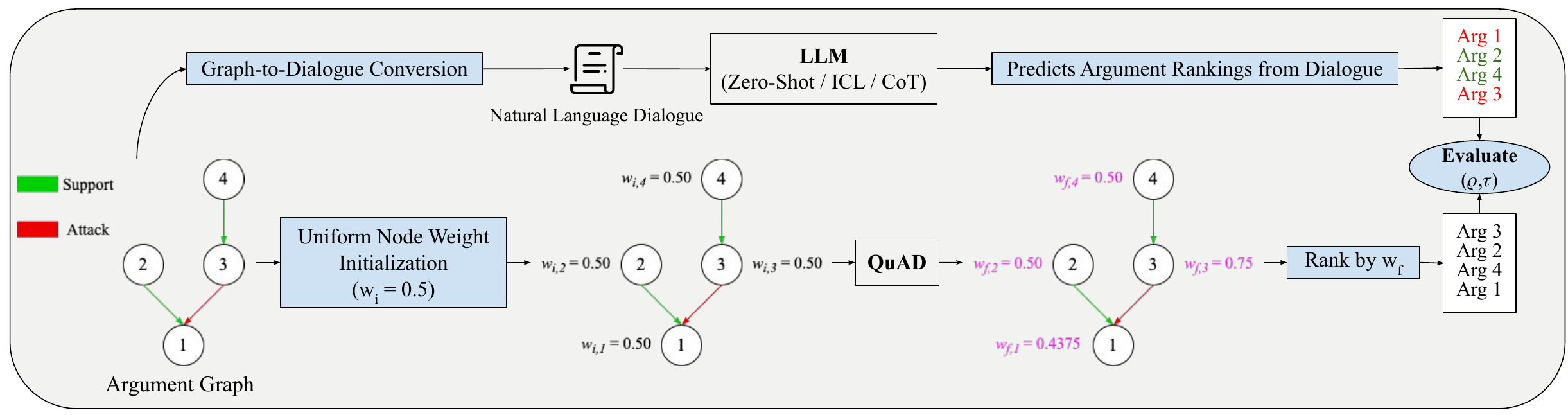}
    \caption{
        System overview showing the LLM pipeline (top), which is compared with QuAD semantics (bottom). 
        Top: We transform the argument graph (unweighted) into a natural language dialogue, which is fed to an LLM with various instruction strategies (zero-shot, ICL, CoT).
        Bottom: 
        After uniform weight initialization, the graph is scored using QuAD semantics, producing final argument strengths $w_f$. These are used to derive a gold ranking, which is compared to rankings predicted by the LLM via Spearman's $\rho$ and Kendall's $\tau$.
    }
    \label{fig:pipeline}
\end{figure*}

Thus, our main research question is to investigate {\em whether contemporary LLMs can effectively reason over non-linear argumentative structures inherent in natural debates}.

To answer this question, we design a novel evaluation setting: we transform debate graphs into natural dialogues, creating \textit{latent argument graphs}—simple argument lists without explicit attack or support relationships. This flattened dialogue format mimics how individuals interact in real-world group discussions, where argument structure is not overtly marked. We instruct the LLM to rank arguments by their inferred acceptability degrees. The model-generated rankings are then compared against ground-truth acceptability degrees computed using QuAD semantics. Figure~\ref{fig:pipeline} summarizes our evaluation pipeline.

The main contributions of this paper are:

\begin{compactenum}
    \item The first systematic evaluation of LLMs' abilities to reason over non-linear argument structures within realistic debate settings, benchmarked against CAT semantics (QuAD).
    \item A characterization of the limits of LLM reasoning on argument ranking through advanced instruction techniques (In-Context Learning and Chain-of-Thought).
    \item An exploration of how debate complexity, discourse order, argument length, and model size impact performance, uncovering strong chronological and structural biases. While advanced instruction methods partially mitigate these biases, even larger models consistently struggle, highlighting the need for future graph-aware language modeling.
\end{compactenum}

Our findings highlight both the potential and the limitations of LLMs as impartial moderators in complex, multi-agent argumentative scenarios and lay crucial groundwork toward more robust reasoning capabilities.

\section{Related Work}

    \subsection{CAT Semantics}
    CAT models debates as structured graphs, where arguments form nodes and directed edges represent \textit{attack} or \textit{support} relationships between arguments. CAT semantics define how to compute each argument’s \textit{acceptability degree}, a numerical value that reflects its strength within the debate structure.

    A prominent class of CAT semantics, known as \textit{bipolar gradual semantics}, includes QuAD~\cite{baroni2018handbook}, DF-QuAD~\cite{rago2016discontinuity}, exponent-based semantics~\cite{amgoud2018evaluation}, and the Quadratic Energy Model~\cite{potyka2018continuous}. These share two core features:

    \begin{compactdesc}
    \item[\textbf{Bipolarity:}] Both attack and support relations influence an argument’s score.
    \item[\textbf{Graduality:}] Arguments receive continuous acceptability degrees rather than binary labels,
    allowing nuanced distinctions between them.
    \end{compactdesc}

    We use QuAD semantics to generate gold-standard rankings for evaluation because it (a) is widely used in the CAT literature, (b) intuitively balances attack and support dynamics to capture the nuanced interplay of arguments, (c) reliably converges on acyclic graphs (the class we study), and (d) is grounded in principles that have been shown to align closely with human judgments of argumentative strength~\citep{vesic2023graphical}. Importantly, this grounding makes QuAD a natural proxy for assessing how well models capture human-like reasoning. A detailed explanation of QuAD and its recursive scoring method is provided in Appendix~\ref{sec:appendix-b}.

    \subsection{CAT and LLMs}

    LLMs' success on diverse reasoning tasks makes them compelling candidates for automated decision-making~\cite{ouyang-li-2023-autoplan}. Yet, they exhibit limitations such as hallucinations and logical inconsistencies~\cite{berglund2024the, fluri2024evaluating}. These deficiencies and a lack of explainability raise concerns about their trustworthiness~\cite{10.1007/s00146-021-01251-8}, motivating research into more structured reasoning frameworks.

    CAT has gained traction for its structured, formal approach to reasoning, and it offers a promising foundation for work on non-linear inference. \citet{castagna2024formal} propose MQArgEng, a pipeline that uses a computational argumentation engine to guide LLM outputs. Their results show that this integration improves reasoning quality without model architecture constraints. \citet{freedman2025argumentative} introduce argumentative LLMs, which construct argumentation frameworks from model-generated reasoning traces and analyze them with gradual semantics to enhance explainability.

    These studies focus on using CAT to enhance LLM outputs by building explicit argumentation graphs. In contrast, we assess whether LLMs can perform non-linear reasoning by implicitly modeling structured argumentation semantics—without access to the underlying graph—when ranking arguments in natural debates.

    To our knowledge, this is the first systematic evaluation of LLMs' ability to capture \textit{gradual, bipolar argumentation semantics} from dialogue alone.

\section{Data}

    \subsection{Dataset Overview and Characteristics}

     We compile our datasets from two sources within the NoDE benchmark~\cite{cabrio:hal-01095073}: \textit{12AngryMen}, derived from a well-known jury deliberation play, and \textit{DebatePedia}, which covers multiple smaller debates on diverse topics. Using these complementary datasets—one large and domain-specific, the other smaller and topic-diverse—we comprehensively assess LLM performance on debates of varying size, complexity, and subject matter. All graphs are acyclic. 
     Details of the datasets are provided below; additional descriptions and examples are included in Appendix~\ref{sec:appendix-a}.

    \paragraph{12AngryMen Dataset:}
    This dataset, based on the play \textit{Twelve Angry Men}, includes three acts represented as argument graphs, each with arguments as nodes and relations (attacks/supports) as edges. The acts contain 39, 33, and 11 nodes, respectively, totaling 80 edges, and are all used for testing.

    \paragraph{DebatePedia Dataset:}
    This dataset comprises debates manually curated from encyclopedias of pro and con arguments. Each debate forms a separate argument graph with user-generated arguments linked by attack or support edges. Chronological ordering preserves dialogue structure. From the total of 22 debates, we select three diverse examples (attack-heavy, support-heavy, and balanced) for instruction exemplars. The remaining 19 debates, averaging 13 nodes each (242 nodes, 223 edges), serve for evaluation.

    \paragraph{Evaluation scale.}
    Across our two datasets we have 325 arguments linked by 303 relations. For evaluation, models must reason over all unordered pairs of arguments within each graph, i.e., $\sum_g \binom{|A_g|}{2} \approx 3{,}000$ pairs, where $|A_g|$ is the number of arguments in graph $g$. With 6 instruction strategies and 3 repetitions, this expands to nearly 50{,}000 pairwise checks per model, underscoring the substantive reasoning workload.

    \subsection{From Argument Graphs to Dialogues}
    To evaluate models in a realistic dialogue-based setting, we convert argument graphs into dialogues. Specifically, we preserve only the arguments' texts, discarding their relationships (attacks or supports). Arguments are chronologically integrated into debates, formatted simply as ``Argument \#: Argument Text''. This conversion preserves the natural discourse flow but keeps argument structure latent, requiring models to implicitly infer relationships. Refer to Appendix~\ref{app:graph-to-dialogue} for more details.

    \subsection{Argument Ranking Using QuAD}

    To evaluate whether LLMs can reason effectively about argument strength in debates, we compare their predicted rankings to those induced by QuAD semantics~\cite{baroni2018handbook}.

    QuAD assigns each argument an \textit{acceptability degree} by considering both its initial weight and the influence of arguments that attack or support it. Supporters increase an argument’s acceptability, while attackers decrease it, and the final score reflects a balance between these competing influences. Importantly, the strength of each attacker or supporter depends recursively on how well-supported \textit{they} are—capturing the idea that a weak rebuttal should count less than a strong one.

    This process resembles a kind of recursive influence propagation: each argument's strength is updated based on the strengths of its neighbors, much like a random walk spreading activation across the argument graph. Over multiple iterations, the system converges to stable values that reflect the global argumentative structure.

    While QuAD produces numerical acceptability scores, we converted these to rankings by sorting arguments in descending order of their final scores. These rankings serve as a gold standard against which LLM-generated rankings are evaluated. The models never observe the underlying graph or edge types—their task is to rank arguments in natural debates in a way that approximates this structured, graph-driven reasoning process.

    It is important to note that QuAD and its derived semantics are defined only for acyclic graphs. We recognize the need for further exploration of cyclic graph structures and address this in Section~\ref{Limitations}.

    Finally, since our datasets do not contain predefined initial argument weights, we assign all arguments a uniform weight of 0.5. This ensures a neutral and consistent evaluation, allowing LLMs to approach each argument as equally important at the outset—effectively starting from a “blank slate” without external bias.

    Formal definitions and detailed examples of QuAD semantics are available in Appendix~\ref{sec:appendix-b}.

\section{Approach}        
    
   \subsection{Models}
    We use four LLMs—both open and closed—GPT-4o~\cite{openai2024gpt4ocard}, Claude~3~Sonnet~\cite{TheC3}, Command~R+~\cite{cohere2024commandr}, and Llama~3~70B~Instruct~\cite{Dubey2024TheL3}—abbreviated as gpt-4o, claude-3, cmd-r-plus, and llama-3. Under identical instruction sets (no fine-tuning) and a temperature of 0.7, we run each model three times and report averaged results.

    \subsection{Instruction Strategies} \label{prompting-approaches}
     Our experiments evaluate LLMs' non-linear reasoning ability by measuring their performance in replicating formal QuAD semantics from latent graphs formatted as debate dialogues. 
    To this end, we explored state-of-the-art instruction strategies, such as Chain-of-Thought (CoT) and In-Context Learning (ICL). For examples, see Appendix~\ref{sec:appendix-c}.

    \paragraph{Zero-Shot ``Vanilla'':} In this method, the LLM is instructed to rank arguments based on their logical strength and their (latent) attack or support relations with other arguments.
  
    \paragraph{In-Context Learning (ICL):} ICL uses exemplars to improve LLM performance~\cite{NEURIPS2020_1457c0d6}. We selected three argument graphs from DebatePedia as ICL examples, covering attack-heavy, balanced, and support-heavy structures. We tested two variants: \textbf{one-shot ICL} (single exemplar) and \textbf{few-shot ICL} (three exemplars).
        
    \paragraph{Zero-Shot Chain-of-Thought (CoT):} Inspired by the zero-shot CoT framework~\cite{NEURIPS2022_8bb0d291}, this approach encourages the LLM to simulate a step-by-step reasoning process about each argument's logical strength and its relationships with other arguments. The model then constructs an adjacency list representing these relationships and ranks the arguments based on their logical strength and the attack or support relationships.

    \paragraph{Chain-of-Thought (CoT):} This approach involves adding detailed analyses of attack and support relationships between arguments and explicitly constructing adjacency lists for each ICL exemplar. This process teaches models to adopt a step-by-step reasoning process when analyzing arguments. We test two variants: \textbf{one-shot CoT} (single exemplar) and \textbf{few-shot CoT} (three exemplars).
  
    \subsection{Debate Structure Reconstruction} \label{sec:graph-recovery}
    Beyond ranking, we evaluate whether models can \emph{reconstruct} the underlying argument graph from dialogue alone. Under our CoT settings, models are instructed to (i) infer pairwise relations and (ii) emit a signed, directed adjacency list as keyed lists, e.g., 
    \texttt{'Argument 2': [('Argument 6','attack'), ('Argument 9','support')], ...}, 
     where each tuple encodes an edge $(i \leftarrow j, r)$ with $r \in \{\textsc{attack}, \textsc{support}\}$.
        
\section{Results}

As mentioned, our experiments evaluate the non-linear reasoning ability of LLMs by assessing their capability to produce similar argument rankings as the ones generated by QuAD semantics. Notably, QuAD semantics has full access to the argument graph structure, whereas the LLMs operate in a more natural dialogue setting. We evaluate the correlation between the LLM rankings and those derived from QuAD using two metrics: (a) Kendall’s $\tau$, which measures directional agreement based on pairwise orderings, and (b) Spearman’s $\rho$, which captures the monotonic relationship and is more sensitive to the magnitude of rank differences. Using both metrics allows us to assess the agreement in order and the severity of misalignment. In addition to ranking, we evaluate whether models can \emph{reconstruct} the signed, directed interaction graph from dialogue by comparing the predicted edge set $\hat{E}$ against the gold edge set $E$.
A prediction is correct only if both endpoints and the relation type match. 
We report precision, recall, and F1 at the graph level, macro-averaged within each split.
Our experiments yield the following findings:

\begin{table}
    \centering
    \small 
    \begin{tabular}{l@{}l@{\hspace{0.05in}} l c@{} r@{}l r@{}l }
        \toprule
    Dataset && Technique && \multicolumn{1}{c}{$\rho$} && \multicolumn{1}{c}{$\tau$} \\ \midrule
\multirow{6}{*}{DebatePedia}       &&   Vanilla                && 0.11          && 0.09          \\
        &&   ICL One-Shot           && 0.26          && 0.22          \\
        &&   ICL Few-Shot           && 0.29          && 0.25          \\ 
        &&   CoT Zero-Shot        && 0.32          && 0.27          \\
        &&   CoT One-Shot      && 0.31          && 0.29          \\
        &&   \textbf{CoT Few-Shot  }    && \textbf{0.46}          && \textbf{0.42}          \\

         \bottomrule
    \end{tabular}
    \caption{Average argument ranking performance of four LLMs on DebatePedia, using different instruction strategies. We report Spearman's $\rho$ and Kendall's $\tau$. Bold marks the best-performing technique. All methods outperform the vanilla baseline, with CoT few-shot scoring the highest on average.}
    \label{tab:dpaverages}
\end{table}

\begin{table}[t]
    \centering
    \small 
    \begin{tabular}{cc@{\hspace{0.05in}} r@{}l l c@{} r@{}l r@{}l }
        \toprule
    Act && \#Arg. && Technique && \multicolumn{1}{c}{$\rho$} && \multicolumn{1}{c}{$\tau$} \\ \midrule
\multirow{6}{*}{1}       &&    \multirow{6}{*}{39}  &&   Vanilla                && 0.10          && 0.07          \\
        &&         &&   ICL One-Shot           && 0.19         && 0.14          \\
        &&         &&   ICL Few-Shot           && 0.22          && 0.16          \\ 
        &&         &&   CoT Zero-Shot       && 0.25          && 0.18          \\
        &&         &&   CoT One-Shot    && 0.26          && 0.20          \\
        &&         &&   \textbf{CoT Few-Shot}    && \textbf{0.33}            && \textbf{0.25}          \\ \midrule
\multirow{6}{*}{2}       &&     \multirow{6}{*}{33}  &&   Vanilla               && -0.02          && -0.02          \\
        &&         &&   ICL One-Shot           && 0.15          && 0.10          \\
        &&         &&   ICL Few-Shot           && 0.13          && 0.10          \\ 
        &&         &&   CoT Zero-Shot        && 0.25          && 0.18          \\
        &&         &&   CoT One-Shot      && 0.26          && 0.19         \\
        &&         &&   \textbf{CoT Few-Shot  }    && \textbf{0.28}          && \textbf{0.19}          \\ \midrule
\multirow{6}{*}{3}       &&     \multirow{6}{*}{11}  &&   Vanilla                && 0.25          && 0.19          \\
        &&         &&   ICL One-Shot           && 0.37          && 0.26          \\
        &&         &&   ICL Few-Shot           && 0.36          && 0.26          \\ 
        &&         &&   \textbf{CoT Zero-Shot}       && \textbf{0.43}          && \textbf{0.35}         \\
        &&         &&   \textbf{CoT One-Shot }    && \textbf{0.43}          && \textbf{0.35}          \\
        &&         &&   \textbf{CoT Few-Shot }    && \textbf{0.43}          && \textbf{0.35}          \\ \midrule
\multicolumn{4}{c}{\multirow{6}{*}{Average}}    &   Vanilla              && 0.11          && 0.08          \\ 
        &&         &&   ICL One-Shot           && 0.24          && 0.17          \\
        &&         &&   ICL Few-Shot           && 0.24          && 0.17          \\ 
        &&         &&   CoT Zero-Shot       && 0.31          && 0.24          \\
        &&         &&   CoT One-Shot     && 0.32          && 0.25          \\
        &&         &&   \textbf{CoT Few-Shot }    && \textbf{0.35}          && \textbf{0.26}          \\

         \bottomrule
    \end{tabular}
    \caption{Average argument ranking performance of LLMs across the three acts and overall on 12AngryMen, using different instruction strategies. We report Spearman's $\rho$ and Kendall's $\tau$, bolding the best-performing technique per act or overall. As in Table~\ref{tab:dpaverages}, all methods improve on the baseline; CoT few-shot is top overall. Stronger results on Act 3, similar in size to DebatePedia, suggest debate length affects performance.}
    \label{tab:12angrymenaverages}
\end{table}

\begin{table}[t]
    \centering
    \small 
    \setlength{\tabcolsep}{0.03in}
    \begin{tabular}{ll rrl@{\hspace{0.04in}} rr}
        \toprule
    \multirow[c]{2}{*}{Model} & \multirow[c]{2}{*}{Technique} & \multicolumn{2}{c}{DebatePedia} && \multicolumn{2}{c}{12AngryMen} \\ \cmidrule{3-4} \cmidrule{6-7}
    && \multicolumn{1}{c}{$\rho$} & \multicolumn{1}{c}{$\tau$} && \multicolumn{1}{c}{$\rho$} & \multicolumn{1}{c}{$\tau$} \\ \midrule 
    \multirow{6}{*}{gpt-4o} & Vanilla & 0.15 & 0.13 && 0.14 & 0.10 \\ 
    & ICL One-Shot & 0.28 & 0.24 && 0.24 & 0.17 \\ 
    & ICL Few-Shot & 0.30 & 0.26 && 0.22 & 0.17 \\ 
    & CoT Zero-Shot & 0.36 & 0.30 && 0.31 & 0.23 \\ 
    & CoT One-Shot & 0.33 & 0.31 && 0.33 & 0.25 \\ 
    & CoT Few-Shot & \textbf{0.54} & \textbf{0.50} && \textbf{0.33} & \textbf{0.27} \\ \midrule 
    \multirow{6}{*}{claude-3} & Vanilla & 0.19 & 0.14 && 0.08 & 0.08\\
    & ICL One-Shot & 0.23 & 0.20 && 0.26 & 0.17\\ 
    & ICL Few-Shot & 0.26 & 0.20 && 0.20 & 0.15\\ 
    & CoT Zero-Shot & 0.21 & 0.16 && 0.29 & 0.21\\ 
    & CoT One-Shot & 0.29 & 0.29 && \textbf{0.43} & \textbf{0.32}\\ 
    & CoT Few-Shot & \textbf{0.43} & \textbf{0.39} && 0.36 & 0.27\\ 
    \midrule \multirow{6}{*}{cmd-r-plus} & Vanilla & 0.07 & 0.05 && 0.12 & 0.09\\
    & ICL One-Shot & 0.37 & 0.31 && 0.30 & 0.23\\
    & ICL Few-Shot & 0.32 & 0.28 && 0.26 & 0.19\\
    & CoT Zero-Shot & \textbf{0.47} & \textbf{0.40} && 0.27 & 0.19\\ 
    & CoT One-Shot & 0.30 & 0.28 && 0.27 & 0.22\\ 
    & CoT Few-Shot & 0.43 & 0.37 && \textbf{0.34} & \textbf{0.25}\\ \midrule 
    \multirow{6}{*}{llama-3} & Vanilla & 0.04 & 0.03 && 0.09 & 0.05\\ 
    & ICL One-Shot & 0.16 & 0.12 && 0.15 & 0.10\\ 
    & ICL Few-Shot & 0.29 & 0.25 && 0.25 & 0.17\\ 
    & CoT Zero-Shot & 0.24 & 0.20 && \textbf{0.38} & \textbf{0.30}\\ 
    & CoT One-Shot & 0.32 & 0.29 && 0.23 & 0.19\\ 
    & CoT Few-Shot & \textbf{0.45} & \textbf{0.40} && 0.35 & 0.25\\ 
    \bottomrule 
    \end{tabular}
    \caption{Argument ranking performance of LLMs on 12AngryMen and DebatePedia, using different instruction strategies. We report Spearman's $\rho$ and Kendall's $\tau$; bold entries indicate the best result per model.}
    \label{tab:debate_models}
\end{table}

\begin{table}[h!]
    \centering
    \small
    \setlength{\tabcolsep}{0.08in}
    \begin{tabular}{l c@{\hspace{.01in}}     r@{}l r@{}l    r@{}l  r@{}l   r@{}l r@{}}
    \toprule
    && \multicolumn{3}{c}{Act 1}  && \multicolumn{3}{c}{Act 2} && \multicolumn{3}{c}{Act 3}  \\ \cmidrule{3-5} \cmidrule{7-9} \cmidrule{11-13}
    Model & & \multicolumn{1}{c}{$\rho$} && \multicolumn{1}{c}{$\tau$} && \multicolumn{1}{c}{$\rho$} && \multicolumn{1}{c}{$\tau$} &&  \multicolumn{1}{c}{$\rho$} &&  \multicolumn{1}{c}{$\tau$} \\ \midrule
    \rowcolor{Gray}  \multicolumn{13}{c}{Vanilla} \\
    gpt-4o        && -0.11  && -0.07  && \textbf{0.19} && \textbf{0.11} && 0.35  && \textbf{0.27} \\
    claude-3       && \textbf{0.26}  && \textbf{0.18}  && -0.16 && -0.11 && 0.15  && 0.16  \\
    cmd-r-plus           && 0.12  && 0.08  && -0.12 && -0.08 && \textbf{0.36}  && \textbf{0.27}  \\
    llama-3 && 0.12  && 0.09  && 0.02 && 0.01 && 0.13  && 0.05  \\ 
    \rowcolor{Gray}  
    \multicolumn{13}{c}{ICL One-Shot} \\
    gpt-4o       && -0.06  && -0.02 && \textbf{0.21} && 0.14 && \textbf{0.56}  && \textbf{0.38}  \\
    claude-3       && \textbf{0.46}  && \textbf{0.33}  && -0.01 && -0.03 && 0.32  && 0.20  \\
    cmd-r-plus           && 0.24  && 0.16  && \textbf{0.21} && \textbf{0.16} && 0.46  && \textbf{0.38}  \\
    llama-3 && 0.12  && 0.09  && 0.19 && 0.13 && 0.15  && 0.09  \\ 
    \rowcolor{Gray}  
    \multicolumn{13}{c}{ICL Few-Shot} \\
    gpt-4o        && -0.06  && -0.02  && 0.21 && 0.14 && \textbf{0.51}  && \textbf{0.38}  \\
    claude-3       && \textbf{0.43}  && \textbf{0.30}  && -0.08 && -0.05 && 0.25  && 0.20  \\
    cmd-r-plus           && 0.24  && 0.16  && 0.04 && 0.05 && \textbf{0.51}  && 0.35  \\
    llama-3 && 0.26  && 0.19  && \textbf{0.34} && \textbf{0.24} && 0.15  && 0.09  \\ 
    \rowcolor{Gray}  
    \multicolumn{13}{c}{CoT Zero-Shot} \\
    gpt-4o        && 0.11  && 0.09  && 0.30 && 0.18 && 0.51  && 0.42  \\
    claude-3       && \textbf{0.47}  && \textbf{0.33}  && 0.17 && 0.14 && 0.22  && 0.16  \\
    cmd-r-plus           && 0.20  && 0.13  && 0.23 && 0.14 && 0.39  && 0.31  \\
    llama-3 && 0.21  && 0.15  && \textbf{0.31} && \textbf{0.25} && \textbf{0.61}  && \textbf{0.49}  \\ 
    \rowcolor{Gray}  
    \multicolumn{13}{c}{CoT One-Shot} \\
    gpt-4o       && 0.30  && \textbf{0.23}  && 0.22 && 0.17 && 0.47  && 0.35  \\
    claude-3       && \textbf{0.33}  && 0.22  && \textbf{0.40} && \textbf{0.28} && \textbf{0.55}  && \textbf{0.45}  \\
    cmd-r-plus           && 0.23  && 0.17  && 0.26 && 0.21 && 0.32  && 0.27  \\
    llama-3 && 0.19  && 0.16  && 0.14 && 0.11 && 0.36  && 0.31  \\ 
    \rowcolor{Gray}  
    \multicolumn{13}{c}{CoT Few-Shot} \\
    gpt-4o        && 0.30  && 0.24  && \textbf{0.29} && \textbf{0.22} && 0.41  && 0.35  \\
    claude-3       && 0.30  && 0.23  &&  \textbf{0.29} && 0.19 && \textbf{0.49} && \textbf{0.38} \\
    cmd-r-plus           && 0.33  && 0.22  && 0.27 && 0.19 && 0.41  && 0.35  \\
    llama-3 && \textbf{0.37}  && \textbf{0.29}  && 0.27 && 0.14 && 0.41  && 0.31  \\
    \bottomrule
    \end{tabular}
    \caption{Argument ranking performance of various LLMs on the three acts of 12AngryMen, using different instruction strategies. We report Spearman's $\rho$ and Kendall's $\tau$. Bold marks the best LLM per metric within each act. Although performance varies across models and acts, all instruction strategies improve results. Notably, every model does better on the smaller Act 3, suggesting challenges in larger, more complex debates.}
    \label{tab:12angrymentotal}
\end{table}

\begin{table}[t]
    \centering
    \small
    \setlength{\tabcolsep}{6pt}
    \begin{tabular}{cc l S[table-format=1.2] S[table-format=1.2] S[table-format=1.2]}
    \toprule
    \textbf{Act} & \textbf{\#Args} & \textbf{Technique} & {\textbf{P}} & {\textbf{R}} & {\textbf{F1}} \\
    \midrule
    \multirow{3}{*}{1} & \multirow{3}{*}{39} & CoT Zero-Shot    & 0.41 & 0.43 & 0.41 \\
                       &                      & CoT One-Shot  & 0.55 & 0.56 & \bfseries 0.56 \\
                       &                      & CoT Few-Shot  & 0.41 & 0.53 & 0.45 \\
    \midrule
    \multirow{3}{*}{2} & \multirow{3}{*}{33} & CoT Zero-Shot    & 0.32 & 0.40 & 0.34 \\
                       &                      & CoT One-Shot  & 0.47 & 0.56 & 0.51 \\
                       &                      & CoT Few-Shot  & 0.52 & 0.58 & \bfseries 0.54 \\
    \midrule
    \multirow{3}{*}{3} & \multirow{3}{*}{11} & CoT Zero-Shot    & 0.26 & 0.48 & 0.33 \\
                       &                      & CoT One-Shot  & 0.67 & 0.77 & \bfseries 0.71 \\
                       &                      & CoT Few-Shot  & 0.68 & 0.73 & 0.70 \\
    \midrule
    \multicolumn{2}{c}{}                 & CoT Zero-Shot    & 0.33 & 0.44 & 0.36 \\
    \multicolumn{2}{c}{\textit{Average}} & CoT One-Shot  & 0.56 & 0.63 & \bfseries 0.59 \\
    \multicolumn{2}{c}{}                 & CoT Few-Shot  & 0.54 & 0.61 & 0.56 \\
    \bottomrule
    \end{tabular}
    \caption{Adjacency-list recovery averages on \textit{12AngryMen}. Best F1 per act and overall are bolded. CoT instructions consistently improve graph recovery. Models perform better on the short Act 3.}
    \label{tab:12am-adj}
\end{table}

\begin{table}[t]
        \centering
        \small
        \setlength{\tabcolsep}{6pt}
        \begin{tabular}{c l S[table-format=1.2] S[table-format=1.2] S[table-format=1.2]}
        \toprule
        \textbf{Avg.\ \ \# Args} & \textbf{Technique} & {\textbf{P}} & {\textbf{R}} & {\textbf{F1}} \\
        \midrule
        & CoT Zero-Shot    & 0.24 & 0.48 & 0.31 \\
        13 & CoT One-Shot  & 0.54 & 0.56 & 0.54 \\
        & CoT Few-Shot  & 0.71 & 0.71 & \bfseries 0.71 \\
        \bottomrule
        \end{tabular}
        \caption{Adjacency-list recovery averages on \textit{DebatePedia}. Best F1 overall is bolded. CoT instructions consistently improve graph recovery.}
        \label{tab:debatepedia-adj}
\end{table}

{\flushleft \textbf{LLMs can moderately rank arguments in debate dialogues.}} 
Tables~\ref{tab:dpaverages} and~\ref{tab:12angrymenaverages} summarize the results of our study, showcasing the average performance of four state-of-the-art LLMs across the DebatePedia and 12AngryMen datasets, respectively. Detailed results per LLM are provided in Table \ref{tab:debate_models}.

We show that off-the-shelf LLMs, 
when used with prompting best practices,  
generally have a decent ability to rank arguments based on their latent graphs. 
This is supported by the moderately positive $\tau$ and $\rho$ values across the results. For instance, in the DebatePedia dataset, the highest average $\rho$ achieved is $0.46$ with the CoT few-shot technique, and the corresponding $\tau$ is $0.42$. The 12AngryMen Dataset results show a similar, albeit lower,  performance. 
Now, we address this difference.

{\flushleft  \textbf{LLMs' ability to rank arguments is significantly influenced by input size.}} Table~\ref{tab:12angrymenaverages} breaks down the average performance of the tested LLMs across the three acts of the 12AngryMen dataset. In Acts 1 and 2, which have four and three times more arguments than Act 3, the LLMs have a significant performance drop in correlation metrics compared to Act 3. DebatePedia's debates, on average, are similar in size to Act 3 of 12AngryMen, making the input size of their experiments much smaller than Acts 1 and 2 of 12AngryMen.  The pronounced performance difference of the same instruction methods between the two datasets (e.g., a 24\% decrease in $\rho$ and a 38\% decrease in $\tau$ for CoT few-shot's average performance in 12AngryMen compared to DebatePedia) seen in Tables~\ref{tab:dpaverages} and~\ref{tab:12angrymenaverages} could be explained similarly. Further, comparing the results on individual acts of 12AngryMen and the averages for DebatePedia clearly shows that smaller input sizes consistently lead to better performance.
\begin{table}
    \centering
    \small 
    \setlength{\tabcolsep}{0.045in}
    \begin{tabular}{l@{} l@{\hspace{0.05in}} l c@{} r r@{}l r r@{}l }
        \toprule
\multirow[c]{2}{*}{Dataset} && \multirow[c]{2}{*}{Technique} && \multicolumn{2}{c}{$\rho$} &&  \multicolumn{2}{c}{$\tau$} \\ \cmidrule{5-6} \cmidrule{8-9}
&& && mean & s.d. && mean & s.d. \\ \midrule
\multirow{6}{*}{12AngryMen} &&   Vanilla && 0.11 & 0.08 && 0.07 & 0.07 \\
&&   ICL One-Shot  && 0.13 & 0.08 && 0.09 & 0.05 \\
&&   ICL Few-Shot && 0.16 & 0.09 && 0.11 & 0.06 \\
&&   CoT Zero-Shot   && 0.33 & 0.10 && 0.26 & 0.08 \\
&&   CoT One-Shot  && 0.05 & 0.14 && 0.06 & 0.10 \\
&&   CoT Few-Shot && 0.17 & 0.11 && 0.14 & 0.09 \\ \midrule
\multirow{6}{*}{DebatePedia}       &&   Vanilla && -0.14 & 0.15 && -0.11 & 0.12 \\
&&   ICL One-Shot  && -0.01 & 0.16 && -0.02 & 0.12 \\
&&   ICL Few-Shot  && 0.01 & 0.19 && 0.01 & 0.14 \\
&&   CoT Zero-Shot && 0.06 & 0.19 && 0.04 & 0.15 \\
&&   CoT One-Shot && 0.29 & 0.15 && 0.25 & 0.12 \\
&&   CoT Few-Shot  && 0.34 & 0.13 && 0.30 & 0.11 \\

         \bottomrule
    \end{tabular}
    \caption{Average argument ranking performance of Llama-3-70b-instruct on 12AngryMen and DebatePedia, under different instruction strategies and five random topological sorts. We report means and standard deviations of Spearman's $\rho$ and Kendall's $\tau$. Compared to the original (chronological) setup in Table~\ref{tab:debate_models}, performance drops overall, indicating reliance on dialogue's chronological flow rather than its argument graph structure.}
    \label{tab:topsort_averages}
\end{table}

{\flushleft \textbf{Advanced instruction methods generally enhance LLM performance in understanding and ranking arguments.}} This is evident from the improved correlation metrics in Tables~\ref{tab:dpaverages} and~\ref{tab:12angrymenaverages}. On average, few-shot CoT consistently outperforms other techniques across different datasets and acts, indicating the effectiveness of combining the CoT approach with few-shot ICL for ranking tasks in natural language argument debates. 

{\flushleft \textbf{Models vary in performance but show similar trends.}} While individual LLMs show varying performance levels in ranking arguments, they exhibit similar trends in response to different instruction strategies and input sizes. To confirm our previous observations from the average LLM performance, we analyze the detailed results of our models for the 12AngryMen dataset's three acts in Table~\ref{tab:12angrymentotal}.

\begin{table*}
    \centering
    \small 
    \setlength{\tabcolsep}{0.045in}
    \begin{tabular}{l c@{} rrrr@{}l@{\hspace{0.1in}} rrrr@{}l@{\hspace{0.1in}}  rrrr@{}l@{\hspace{0.1in}} rrrr@{}l}
        \toprule
        && \multicolumn{10}{c}{Argument Length} & \multicolumn{9}{c}{Argument Position} \\ \cmidrule{3-11} \cmidrule{13-21}
 \multirow[c]{2}{*}{Technique} && \multicolumn{4}{c}{$\rho$} &&  \multicolumn{4}{c}{$\tau$}  && \multicolumn{4}{c}{$\rho$} &&  \multicolumn{4}{c}{$\tau$} \\ \cmidrule{3-6} \cmidrule{8-11} \cmidrule{13-16} \cmidrule{18-21}
&& \multicolumn{1}{c}{LQ1} & \multicolumn{1}{c}{LQ2} & \multicolumn{1}{c}{LQ3} & \multicolumn{1}{c}{LQ4} && \multicolumn{1}{c}{LQ1} & \multicolumn{1}{c}{LQ2} & \multicolumn{1}{c}{LQ3} & \multicolumn{1}{c}{LQ4} && \multicolumn{1}{c}{PQ1} & \multicolumn{1}{c}{PQ2} & \multicolumn{1}{c}{PQ3} & \multicolumn{1}{c}{PQ4} && \multicolumn{1}{c}{PQ1} & \multicolumn{1}{c}{PQ2} & \multicolumn{1}{c}{PQ3} & \multicolumn{1}{c}{PQ4} \\ \midrule
 Vanilla && 0.42 & 0.26 & 0.11 & 0.24 && 0.38 & 0.27 & 0.08 & 0.21 && 0.18 & 0.31 & -0.04 & 0.19 && 0.18 & 0.27 & -0.01 & 0.20\\
  ICL One-Shot  && 0.50 & 0.35 & 0.25 & 0.29 && 0.46 & 0.33 & 0.23 & 0.27 && 0.09 & 0.49 & 0.12 & 0.34 && 0.08 & 0.46 & 0.14 & 0.34\\
  ICL Few-Shot && 0.46 & 0.37 & 0.25 & 0.41 && 0.43 & 0.35 & 0.25 & 0.38 && 0.13 & 0.42 & 0.25 & 0.38 && 0.10 & 0.41 & 0.24 & 0.40\\
  CoT Zero-Shot   && 0.31 & 0.27 & 0.30 & 0.16 && 0.31 & 0.27 & 0.27 & 0.14 && 0.17 & 0.41 & 0.19 & 0.22 && 0.16 & 0.41 & 0.18 & 0.24\\
  CoT One-Shot  && 0.68 & 0.52 & 0.31 & 0.54 && 0.66 & 0.51 & 0.29 & 0.53 && 0.51 & 0.63 & 0.56 & 0.39 && 0.43 & 0.64 & 0.55 & 0.41\\
  CoT Few-Shot && 0.68 & 0.64 & 0.69 & 0.63 && 0.67 & 0.65 & 0.67 & 0.61 && 0.41 & 0.72 & 0.67 & 0.51 && 0.29 & 0.71 & 0.67 & 0.53\\ 
    \bottomrule
    \end{tabular}
    \caption{Average argument ranking performance of LLMs on DebatePedia, split into quartiles by argument length and position (L/PQ1--L/PQ4). We report Spearman's $\rho$ and Kendall's $\tau$ under various instruction strategies. The Vanilla approach exhibits length and positional biases, while ICL mitigates length bias but not positional bias. CoT addresses both, maintaining consistently strong performance across quartiles.}
    \label{tab:arg_len_averages}
\end{table*}

Despite architectural differences, all models benefit from advanced methods such as CoT and ICL. Notably, each model improves significantly when shifting from the baseline Vanilla approach to CoT few-shot instructions.

Another noteworthy trend is how input size affects each model similarly. In the substantially smaller Act 3, all models tend to achieve higher correlation metrics. This pattern suggests that models struggle with larger input sizes, likely due to the increased complexity and potential limitations in processing longer contexts.
Despite these overarching trends, individual models exhibit unique performance nuances. For example, Claude 3 Sonnet excels in Act 1 with ICL techniques but shows inconsistent results in Act 2, even producing negative correlations under the same instruction methods. Interestingly, models with relatively decent performance in Vanilla ranking ($\rho > 0.2$) exhibit exceptional results under ICL ($\rho \in [0.43,\,0.56]$). This phenomenon might be related to these models surfacing memorized training data when using ICL. Recent research by~\citet{golchin2024memorizationincontextlearning} suggests that ICL can trigger LLMs to retrieve and utilize memorized data from their training corpus, enhancing performance. 
Consequently, future work on benchmarking these models' reasoning abilities should probe them for data contamination and memorization on the targeted test data.

These variations highlight that while the models follow similar trends, their performance can be influenced by specific interactions between their architectures, training methods, and the evaluation dataset's characteristics. Further, these results suggest that for a complex task such as CAT, architectures based on mixtures of experts might be better and more stable than individual models. This observation is supported by comparing the per-LLM breakdown in Tables~\ref{tab:debate_models} and~\ref{tab:12angrymentotal} to the averages in Tables~\ref{tab:dpaverages} and~\ref{tab:12angrymenaverages}. While no single model consistently dominates across all acts and instruction techniques, the average results provide a more stable and consistently good performance. 

{\flushleft \textbf{Structure recovery mirrors ranking patterns.}}
Tables~\ref{tab:12am-adj} and~\ref{tab:debatepedia-adj} report results for the \emph{intermediate} task of debate graph reconstruction.
On \textit{12AngryMen}, we observe that macro F1 rises from 0.36~(CoT zero-shot) to 0.59~(CoT one-shot) and 0.56~(CoT few-shot), with Act~III reaching 0.71. 
On \textit{DebatePedia}, CoT with exemplars also performs strongly (0.54 one-shot; 0.71 few-shot).
See Appendix~\ref{sec:appendix-d} for per-act and per-model results.

Three clear patterns emerge:

\begin{compactenum}[(i)]
    \item \emph{CoT exemplars markedly improve edge recovery}. Moving from zero-shot to one/few-shot CoT improves F1 by about {\raise.17ex\hbox{$\scriptstyle\sim$}}0.15–0.40 across datasets.

    \item \emph{Shorter debates are easier}. Smaller graphs (e.g., \textit{12AngryMen} Act~III, typical DebatePedia topics) yield higher F1, mirroring the ranking task’s sensitivity to input length.

    \item \emph{Better recovery aligns with better ranking}. The setups that most improve edge prediction (CoT with exemplars) are also those with the highest rank correlations. This suggests that capturing interaction structure helps models approach QuAD’s ordering.
\end{compactenum}

\section{Discussion}

To better understand LLMs' limits in modeling argumentation semantics, we further analyze their performance along four key aspects: discourse chronology's impact, argument length's influence, positional bias in LLM rankings, and model size.
Lastly, we test whether specialization changes behavior by comparing our general-purpose models to a dedicated \emph{reasoner} (DeepSeek R1).

    \subsection{Influence of Discourse Chronology}
    To evaluate the models' reliance on argument order, we employed topological sorting to randomize the sequence of arguments while preserving their attack and support relationships. Out of all possible topological sorts of the debate's graph, we randomly selected five different sorts for each debate. This shuffles the argument order within a dialogue without altering the graph structure, ensuring the models must infer argument strengths based on their relationships, not their positions. We then applied the same evaluation metrics as in our core experiments to assess the impact of this randomization on the models' ranking performance. Due to its consistent performance across the previous experiments that is similar to the average behavior, our model of choice for this experiment was Llama3 70B Instruct.

    {\flushleft \textbf{Topological sorting lowers performance.}}
    Our results in Table~\ref{tab:topsort_averages} show that topological sorting of arguments generally reduces LLM performance in capturing QuAD semantics. This indicates that LLMs rely more on the natural, chronological flow of dialogue than on the underlying argument graph structure. When arguments are presented in the natural order of the dialogue, LLMs effectively use contextual and conversational cues to infer attack and support relationships. However, rearranging the arguments through topological sorting disrupts this flow, making it harder for the models to recognize these connections.

    \subsection{Influence of Argument Length and Position on Performance}

    We first compute QuAD rankings for each DebatePedia debate and have the LLMs rank all arguments. Then, we run two separate analyses: (1) splitting arguments into quartiles based on individual argument length (measured by token count), and (2) splitting them by their position in the debate, based on their chronological appearance. This preserves the original QuAD computation but reveals any biases tied to argument size or placement.

    {\flushleft \textbf{Vanilla instructing suffers from argument length bias.}}
    Table~\ref{tab:arg_len_averages} presents average LLM performance on DebatePedia, categorized by argument length quartiles (LQ1--LQ4, shortest to longest). Using the Vanilla instruction technique, $\rho$ and $\tau$ are highest for the shortest arguments in LQ1 ($\rho = 0.42$, $\tau = 0.38$), decrease in LQ2, reaching the lowest in LQ3 ($\rho = 0.11$, $\tau = 0.08$), and slightly improve in LQ4 ($\rho = 0.24$, $\tau = 0.21$).
    \begin{figure*}[h!]
        \centering
        \begin{subfigure}{0.46\textwidth}
            \centering
            \includegraphics[width=\linewidth]{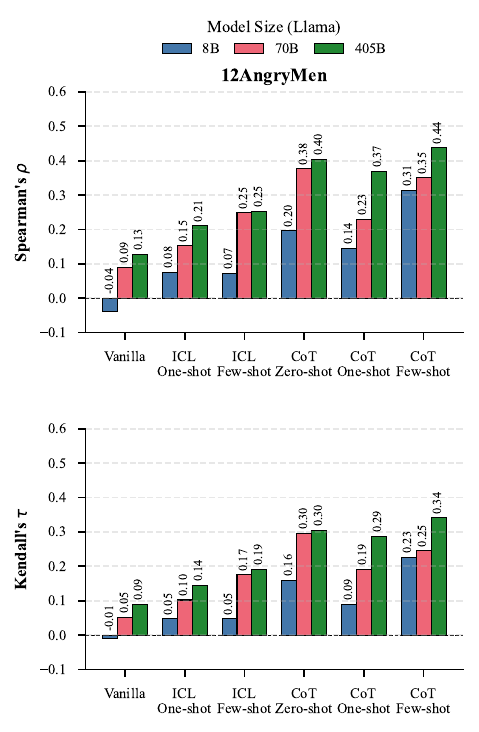}
            \caption{Comparison on 12AngryMen dataset}
            \label{fig:llama_12angrymen}
        \end{subfigure}
        \hspace{2mm}
        \begin{subfigure}{0.46\textwidth}
            \centering
            \includegraphics[width=\linewidth]{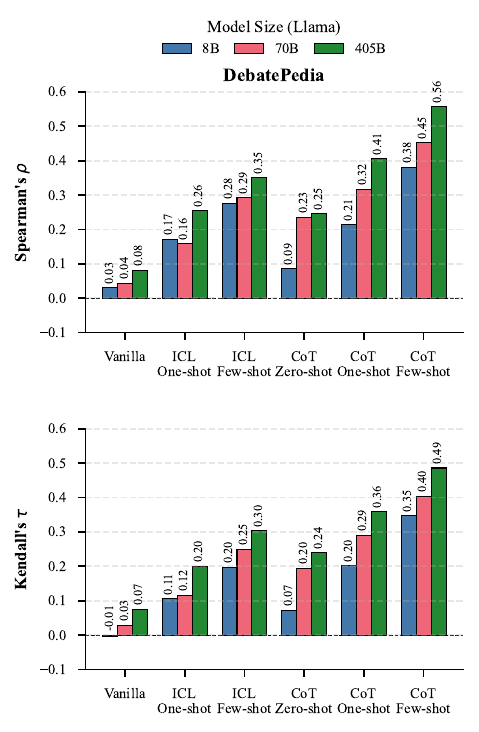}
            \caption{Comparison on DebatePedia dataset}
            \label{fig:llama_debatepedia}
        \end{subfigure}
        \caption{Argument ranking performance comparison of Llama~3~8B, Llama~3~70B, and Llama~3.1~405B across different instruction strategies. Larger models consistently outperform smaller ones, with significant gains in the CoT few-shot setting.}
        \label{fig:llamas}
    \end{figure*}

    These results suggest a length bias in Vanilla instructing, where LLMs perform best on shorter arguments and struggle with longer ones. The slight recovery in LQ4 may indicate that longer arguments offer enough context for better judgment, mitigating some of the drop from LQ3. 
    This bias may arise from models relying on superficial features like length, finding shorter arguments easier to process,
    while medium ones may not provide sufficient context, leading to reduced performance.

    {\flushleft \textbf{In-Context Learning mitigates length bias.}}
    More advanced instruction techniques such as ICL and CoT effectively mitigate the length bias seen in Vanilla instructing. As shown in Table~\ref{tab:arg_len_averages}, CoT few-shot instructions maintain high performance across all argument lengths, with $\rho$ ranging from 0.63 to 0.69 and $\tau$ from 0.61 to 0.67.

    This consistency highlights how reasoning steps and examples help LLMs focus on logical content rather than length, enabling them to handle arguments of varying lengths. Advanced instruction strategies also help the models infer attack and support relationships, aligning more closely with formal argumentation semantics.

    {\flushleft \textbf{Positional bias detected.}}
    Table~\ref{tab:arg_len_averages} also presents the LLMs' average performance by argument position quartiles (PQ1--PQ4, where PQ1 contains the earliest arguments in the discourse and PQ4 the latest). With Vanilla instructions, performance varies across quartiles, revealing a positional bias. $\rho$ peaks in PQ2 ($\rho = 0.31$), is lower in PQ1 ($\rho = 0.18$), turns negative in PQ3 ($\rho = -0.04$), and is moderate in PQ4 ($\rho = 0.19$). $\tau$ follows a similar lead. This pattern, although less pronounced, is evident in the regular ICL instructions too. This suggests LLMs perform better on mid and later arguments (PQ2, PQ4) but struggle with earlier and mid-sequence arguments.

    A possible reason for this bias may be that later arguments benefit from more context, making it easier for LLMs to infer relationships. Consequently, PQ2 and PQ4 outperform the earlier quartiles.

    ICL CoT instruction techniques, e.g., CoT one-shot and few-shot, reduce this bias. For CoT few-shot, $\rho$ ranges from 0.41 to 0.72, with the highest scores in PQ2 and PQ3. $\tau$ also improves, ranging from 0.29 to 0.71. These techniques help LLMs better utilize context in reasoning, making performance more consistent across all quartiles.

    \subsection{Influence of Model Size on Performance}
    In addition to Llama~3~70B, we evaluated Llama~3~8B and Llama~3.1~405B to gauge size effects (Figure~\ref{fig:llamas}). Larger models consistently \textit{outperform} smaller ones, with Llama~3.1~405B achieving the highest correlations in CoT few-shot. Meanwhile, Llama~3~8B struggled on longer debates and often broke formatting constraints. These findings suggest \textit{reasoning ability and instruction fidelity scale with model size}.

    \subsection{Reasoner vs.\ General-Purpose LLMs}
    We evaluate DeepSeek~R1~\cite{deepseekai2025deepseekr1incentivizingreasoningcapability}, a dedicated \emph{reasoner} (685B), under identical setups as the general-purpose models. On \textit{12AngryMen}, R1 is on par with the strongest general-purpose models across settings. On \textit{DebatePedia}, it underperforms \emph{all} general-purpose models. Despite its size, it frequently violates the required output format; one and few-shot CoT reduces format errors but leaves a sizable performance gap (see Appendix~\ref{sec:appendix-e} for per-setting plots vs.\ the 4-model average). 
    These results suggest that specialization for step-by-step reasoning alone does not confer robustness to \emph{non-linear} interactions.

\section{Conclusion}

We investigated whether LLMs can reason over non-linear argumentative structures by ranking arguments in natural debates without access to explicit attack or support relationships. Using CAT semantics—specifically QuAD—as a structured reference, we evaluated LLM performance across multiple instruction strategies.

Our findings show that LLMs can partially approximate structured argumentation reasoning, but their performance is highly sensitive to debate length, argument order, and model size. To our knowledge, this study offers the first systematic analysis of LLMs' ability to recover \textit{gradual bipolar argumentation semantics} from dialogue alone, highlighting both their current limitations and potential for future graph-aware reasoning.

\section*{Limitations} \label{Limitations}

This study has certain constraints. First, QuAD semantics guarantee convergence only on acyclic graphs, so we used only acyclic graphs. Future work will include cyclic graphs where QuAD semantics (or other CAT algorithms) converge.

Second, in all our experiments, we considered all arguments to be equally important initially by assigning them a uniform weight of 0.5 when computing QuAD acceptability degrees. This design ensures a neutral, unbiased evaluation and isolates the model’s ability to infer structure from the dialogue alone. However, in practice, arguments may vary in initial strength based on external knowledge or rhetorical cues. Future work could explore using LLMs themselves to estimate these initial weights based on their parametric memory, prior knowledge, or discourse context.

Third, all the corpora we have worked with in this study are in English, and LLMs' ability to learn gradual bipolar argumentation semantics in other languages might differ. 

Fourth, generalizing claims about LLMs requires the testing of a wide array of models. Given the vast number of LLMs and budget limits, we could only select a fraction of the available options. However, we ensured generalizability by selecting both open-sourced and closed-sourced models that rank highest among different benchmarks.

Lastly, we only used QuAD semantics. While this algorithm is widely used, other bipolar CAT semantics could also serve as valuable benchmarks for LLMs. Future work will explore these additional semantics for a more comprehensive evaluation of LLMs' reasoning capabilities.

\section*{Ethics Statement}

This work uses only publicly available datasets and pretrained language models, without additional training or fine-tuning. No human subjects or personal information are involved. We do not anticipate any ethical concerns arising from this study.

\bibliography{anthology,custom}
\appendix

\section{CAT and QuAD Semantics}
\label{sec:appendix-b}

    \subsection{Overview of CAT} 
    \label{app:cat-overview}
    Imagine a courtroom debate or a lively online discussion thread: arguments are put forward, counter-arguments attack them, and some arguments even bolster others. Rather than a single chain of reasoning, we get a \textit{web of supporting and attacking arguments}, much like a dialogue where evidence and rebuttals interweave. CAT is the field of AI that formalizes such scenarios, enabling us to model and analyze these complex argument networks~\cite{DUNG1995321, baroni2015automatic}. 

    In CAT, an \textit{argumentation framework} is represented as a directed graph where each node is an argument (a claim or proposition) and edges represent \textbf{relations} between arguments. These relations can be of two types: \textbf{attacks}, where one argument challenges or rebuts another, and \textbf{supports}, where one argument provides backing or evidence for another. 

    In our debate analogy, an attorney’s claim might be \textit{attacked} by the opponent’s counter-argument, while a witness’s testimony might \textit{support} the attorney’s claim. CAT thus captures reasoning in a \textbf{non-linear structure}: arguments do not merely follow one another, but branch into pro and con threads that interact in a graph-like manner.

    CAT provides formal tools to determine which arguments ultimately stand (and to what degree) given this relational structure. These tools are known as \textit{argumentation semantics}, which assign each argument an \textbf{acceptability status} based on the full network of attacks and supports.

    While classical semantics might select a subset of “accepted” arguments (as in Dung’s theory,~\cite{DUNG1995321}), \textbf{gradual semantics} instead assign each argument a real-valued \textit{acceptability degree}, often in $[0,1]$. This allows finer-grained distinctions: an argument might be weakly supported or strongly undermined rather than simply accepted or rejected.

    Several such semantics exist. In this work, we focus on the \textbf{QuAD} semantics~\cite{baroni2018handbook}, which belong to the class of \textit{bipolar gradual semantics}—those that consider both attack and support relations and yield a continuous score for each argument. This approach is especially useful for modeling the nuanced interplay of conflicting arguments in natural debates.

    \subsection{Formal Definition of QuAD Semantics}
    \label{app:quad-def}

    We adopt the QuAD semantics~\cite{baroni2018handbook}, which compute argument strengths in a \textbf{quantitative bipolar argumentation framework} (QBAF). Assume an acyclic graph structure.

    Let $\mathcal{F} = \langle A, R^-, R^+, \theta \rangle$ where:
    \begin{itemize}
        \item $A$ is a finite set of arguments;
        \item $R^- \subseteq A \times A$ is the \textbf{attack} relation;
        \item $R^+ \subseteq A \times A$ is the \textbf{support} relation;
        \item $\theta: A \to [0,1]$ assigns an initial \textbf{base weight} to each argument.
    \end{itemize}

    For any $a \in A$, define:
    \begin{align*}
        \mathit{Att}(a) &= \{ b \mid (b,a) \in R^- \} \quad \text{(attackers of $a$)} \\
        \mathit{Sup}(a) &= \{ c \mid (c,a) \in R^+ \} \quad \text{(supporters of $a$)}
    \end{align*}

    The final \textbf{acceptability degree} of $a$, denoted $\sigma(a)$, is defined recursively as:

    {\small
    \begin{equation}
    \label{eq:quad}
    \sigma(a) =
    \begin{cases}
        v_a(a) & \text{if } \mathit{Sup}(a) = \emptyset, \mathit{Att}(a) \neq \emptyset \\
        v_s(a) & \text{if } \mathit{Sup}(a) \neq \emptyset, \mathit{Att}(a) = \emptyset \\
        \theta(a) & \text{if } \mathit{Sup}(a) = \emptyset, \mathit{Att}(a) = \emptyset \\
        \frac{v_a(a) + v_s(a)}{2} & \text{otherwise}
    \end{cases}
    \end{equation}
    }

    Where:
    \begin{align*}
        v_a(a) &= \theta(a) \cdot \prod_{b \in \mathit{Att}(a)} (1 - \sigma(b)) \\
        v_s(a) &= 1 - (1 - \theta(a)) \cdot \prod_{c \in \mathit{Sup}(a)} (1 - \sigma(c))
    \end{align*}

    This definition reflects how strong attackers reduce $\sigma(a)$, while strong supporters increase it. The update rule is evaluated recursively in topological order (as the graph is acyclic), and converges to a unique fixed point for all $a \in A$.

    The resulting $\sigma(a) \in [0,1]$ captures the overall \textit{persuasiveness} or \textit{acceptability} of argument $a$ in the context of the full argumentation structure. Arguments with strong support chains and weak attackers receive high scores; arguments undercut by credible attacks receive lower ones.

    \subsection{Worked Example: SobrietyTest Debate}
    \label{app:quad_example}

    \begin{figure*}[t]
        \centering
        \includegraphics[width=\linewidth]{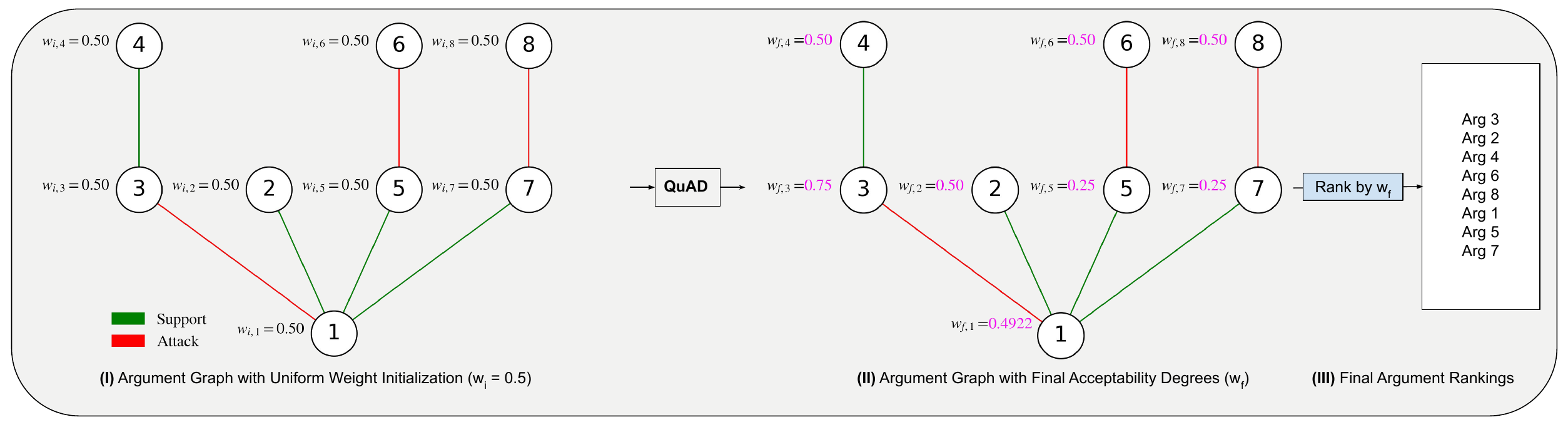}
        \caption{QuAD computation pipeline for the \textit{SobrietyTest} debate. \textbf{(I)} Starting from uniform weight initialization = $\theta(a) = 0.5$ ($w_i$ in graph) on the obtained argument graph (Figure~\ref{fig:base_graph}), \textbf{(II)} we apply QuAD semantics to compute final acceptability degrees $\sigma(a)$ ($w_f$ in graph, pink labels). \textbf{(III)} The resulting ranking of arguments based on their acceptability degrees forms the gold standard used in our evaluation.}
        \label{fig:quad-graph}
    \end{figure*}

    \begin{figure}
        \centering
        \includegraphics[width=0.45\textwidth]{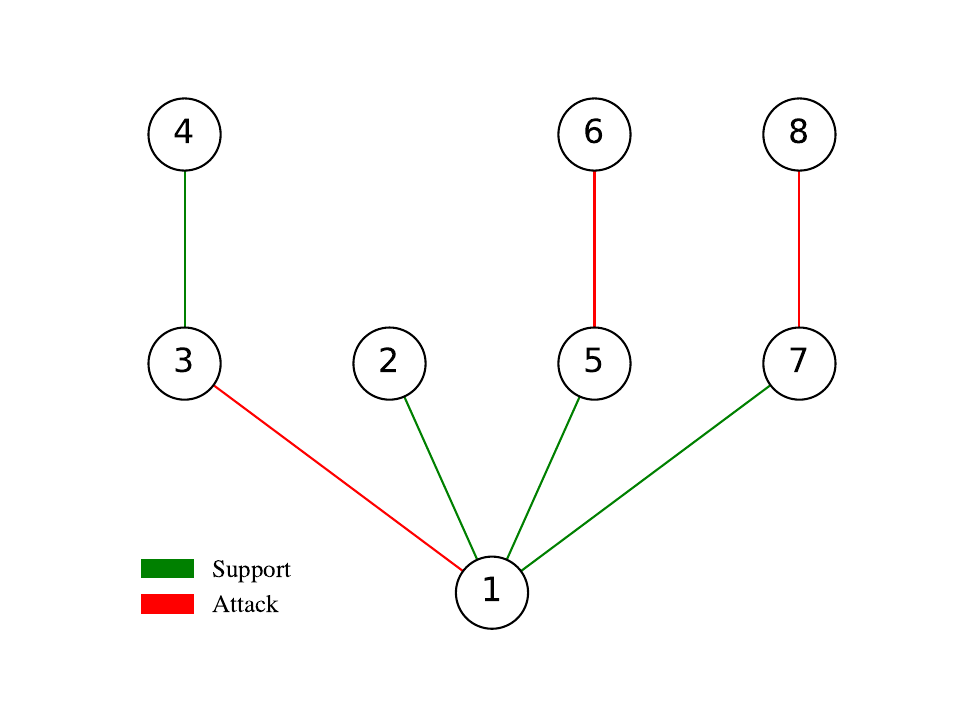}
        \caption{Bipolar argumentation graph for the \textit{SobrietyTest} debate. Nodes represent arguments; red edges indicate attacks and green edges indicate supports.}
        \label{fig:base_graph}
    \end{figure}

    To illustrate how QuAD semantics operate in practice, we walk through an example taken from the \textit{SobrietyTest} debate in the DebatePedia dataset. Figure~\ref{fig:quad-graph} shows the full pipeline used by our system to compute QuAD-based argument rankings.

    The process begins by applying a uniform weight initialization of $\theta(a) = 0.5$ to all arguments in the obtained debate argument graph shown in Figure~\ref{fig:base_graph}. We then apply the QuAD update rule (Equation~\ref{eq:quad}) to compute each argument’s final acceptability degree $\sigma(a)$, propagating influence recursively through the graph. These scores are sorted in descending order to produce the gold-standard QuAD ranking used in our evaluation.

    The argumentation graph contains both support (green edges) and attack (red edges) relations. QuAD semantics capture the intuition that an argument is strengthened by well-supported allies and weakened by strong adversaries.

    For example:
    \begin{itemize}
        \item \textbf{Argument 3} is supported by Argument~4. Because both start with the same base weight ($0.5$), this support lifts Argument~3’s score by 50\%, raising it to about $0.75$.
        \item \textbf{Argument 5} is attacked by Argument~6. With the attacker also at $0.5$, Argument~5 loses half its initial strength, dropping from $0.5$ to $0.25$.
        \item \textbf{Argument 1} faces a \emph{strong attack} from Argument~3, \emph{weak support} from Arguments~5 and~7, and \emph{moderate support} from Argument~2. The net effect nearly cancels the strong attack, so Argument~1’s score decreases only slightly—from $0.5$ to $0.4922$ (a $1.56\%$ reduction).
    \end{itemize}

\section{Dataset Details}
\label{sec:appendix-a}

We use the NoDE benchmark~\cite{cabrio:hal-01095073}, a collection of structured natural language argumentation datasets, to evaluate LLMs’ ability to reason over debates with varying complexity. All datasets contain manually annotated, acyclic bipolar argumentation graphs constructed from pairs of arguments labeled as either \texttt{support} or \texttt{attack}. We use two datasets from NoDE: \textit{DebatePedia} and \textit{Twelve Angry Men}. The third dataset, derived from Wikipedia revision histories, is excluded because it captures edit justifications rather than full debate-style interactions.

    \subsection{NoDE Benchmark Preview}
    \label{app:node-overview}

    Each NoDE dataset contains two annotation layers:
    {\flushleft \textbf{(1)}} Argument pairs labeled with semantic relations.

    {\flushleft \textbf{(2)}} Bipolar argumentation graphs built from those pairs.

    This layered structure supports both the recognition of argumentative relations and the computation of argument strength using CAT semantics. All graphs used in our study are acyclic and annotated with high inter-annotator agreement (Cohen’s $\kappa > 0.7$). A summary of the datasets we use follows.

    \subsection{12AngryMen and DebatePedia Datasets}
    \label{app:node-datasets}

    \paragraph{12AngryMen.}  
    This dataset is built from the dialogue of the play \textit{Twelve Angry Men}, which follows jury deliberations in a homicide trial. Arguments are extracted from the script and linked when one character responds to, supports, or challenges another. Each of the play’s three acts is treated as a separate graph.
    
    \paragraph{DebatePedia.}  
    This dataset includes 22 small debates from DebatePedia and ProCon, platforms focused on user-contributed pro and con arguments. Each debate is transformed into a graph by linking user-submitted arguments based on entailment and contradiction annotations. The original chronological ordering is preserved to maintain a dialogue-like flow.

    \paragraph{Graph structure.}

    \begin{table}
        \centering
        \begin{tabular}{l@{\hspace{0.35in}} r}
        \toprule
        \textbf{Property} & \textbf{Value} \\
        \midrule
        \# Graphs & 3\\
        ~~~~\% With Fan-In & 100\% \\ \midrule
        \# Nodes & 83 \\
        ~~~~\# Act I & 39 \\
        ~~~~\# Act II & 33 \\
        ~~~~\# Act III & 11 \\ 
        Avg.\ In-Degree & 0.96 $\pm$ 1.39 \\
        Avg.\ Out-Degree & 0.96 $\pm$ 0.16 \\\midrule
        \# Edges & 80 \\ 
        ~~~~\# Support Edges & 25 \\
        ~~~~\# Attack Edges & 55 \\ \midrule \midrule
        Agreement (Cohen's $\kappa$) & 0.74 \\
        \bottomrule
        \end{tabular}
        \caption{Summary of the 12AngryMen dataset. All three acts contain convergent structures (\emph{fan-in}), where multiple arguments attack or support the same claim.}
        \label{tab:12angrymen_stats}
        \end{table}

        \begin{table}
            \centering
            \begin{tabular}{l@{\hspace{0.35in}} r}
            \toprule
            \textbf{Property} & \textbf{Value} \\
            \midrule
            \# Graphs & 22 \\
            ~~~~\% With Fan-In & 100\% \\ \midrule
            \# Nodes & 282 \\
            Avg.\ In-Degree & 0.92 $\pm$ 2.14 \\
            Avg.\ Out-Degree & 0.92 $\pm$ 0.25 \\ \midrule
            \# Edges & 260 \\
            ~~~~\# Support Edges & 140 \\
            ~~~~\# Attack Edges & 120 \\ \midrule \midrule
            Agreement (Cohen's $\kappa$) & 0.70 \\ \bottomrule
            \end{tabular}
            \caption{Summary of the DebatePedia dataset. All graphs contain convergent structures (\emph{fan-in}), where multiple arguments attack or support the same claim.}
            \label{tab:debatepedia_stats}
            \end{table}

    As summarized in Tables~\ref{tab:12angrymen_stats}–\ref{tab:debatepedia_stats}, both datasets consistently exhibit convergent structure (\emph{fan-in}): every graph contains at least one argument with in-degree $\geq 2$ (max in-degree ranges 3–11), with mean in-degree near 1 but substantial dispersion (std.\ $\approx$1.4–2.1). By contrast, fan-out is not present (max out-degree = 1 in all graphs). Thus, the non-linearity in these debates arises from multiple premises jointly targeting the same claim—going beyond near-path chains and requiring aggregation of several (often conflicting) influences on a single node.

    \subsection{Graph-to-Dialogue Conversion Example}
    \label{app:graph-to-dialogue}

    To evaluate LLMs in a realistic language setting, we convert structured argumentation graphs into dialogue-style inputs. This transformation preserves argument texts and order but omits attack and support edges. The resulting format mirrors the natural flow of conversation while keeping the graph structure latent.

\begin{figure*}
        \centering
        \includegraphics[width=0.95\textwidth]{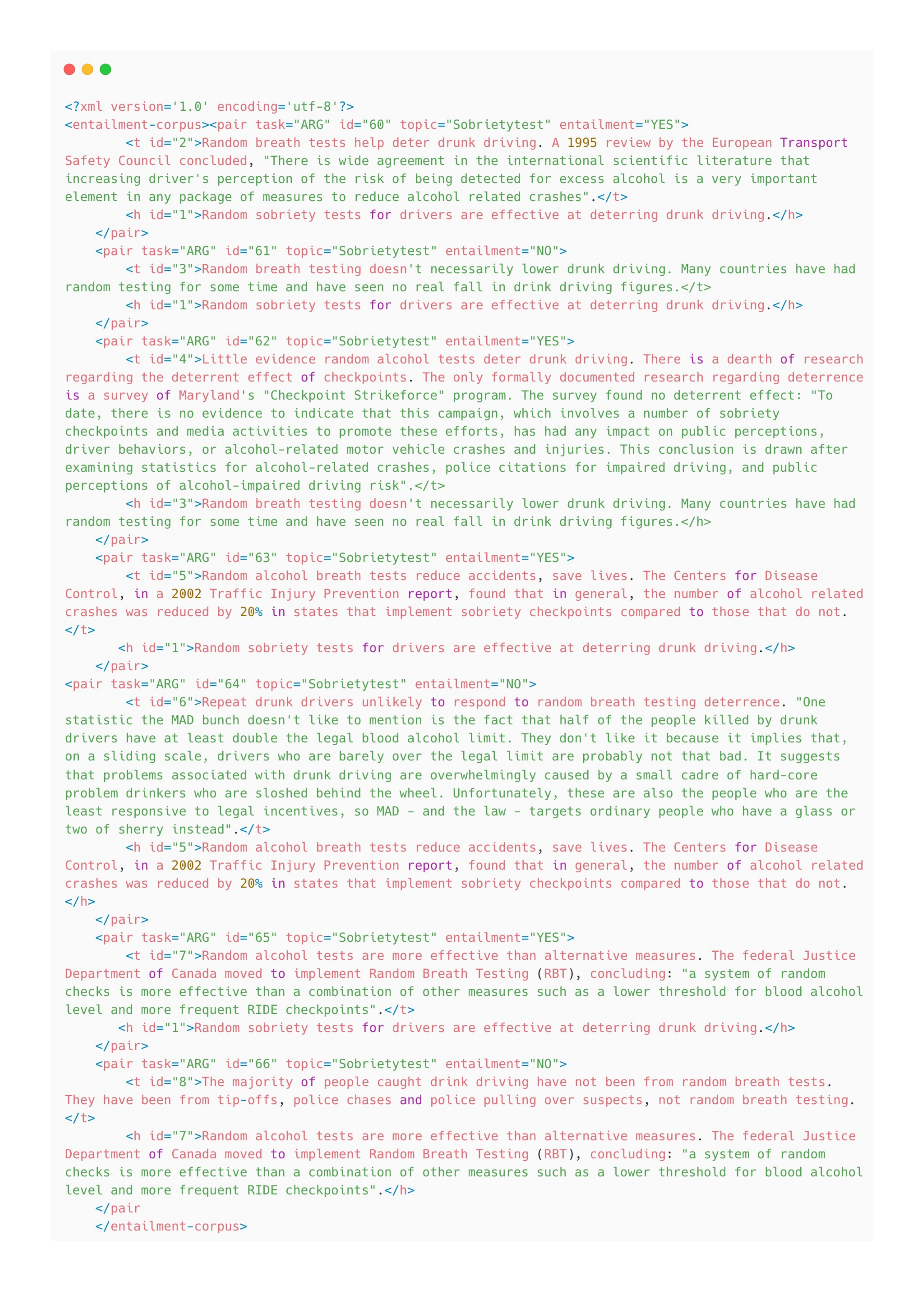}
        \caption{Full XML representation of DebatePedia's \textit{SobrietyTest} debate. Each \texttt{<pair>} contains a claim (\texttt{h}) and its supporting or opposing argument (\texttt{t}) with a labeled entailment relation.}
        \label{fig:xml}
    \end{figure*}

    Figure~\ref{fig:xml} shows the XML source of the \textit{SobrietyTest} debate in the DebatePedia dataset. Each \texttt{pair} element links two arguments using \texttt{text} (\texttt{t}) and \texttt{hypothesis} (\texttt{h}) IDs, with the \texttt{entailment} attribute indicating whether the relation is support (\texttt{YES}) or attack (\texttt{NO}). The \texttt{topic} attribute identifies the debate topic, and the \texttt{id} field gives each pair a unique identifier. Argument IDs are scoped per graph.

    Using the annotated pairs, we construct a bipolar argumentation graph such as the one shown in Figure~\ref{fig:base_graph}, with edges labeled according to their entailment types (support/attack).

    To create the LLM input, we discard all structural information and sort the arguments chronologically by their appearance. Each argument is presented in the format:

    \begin{quote}
    \texttt{Argument \#: Argument Text}
    \end{quote}

    \begin{figure*}
        \centering
        \includegraphics[width=\textwidth]{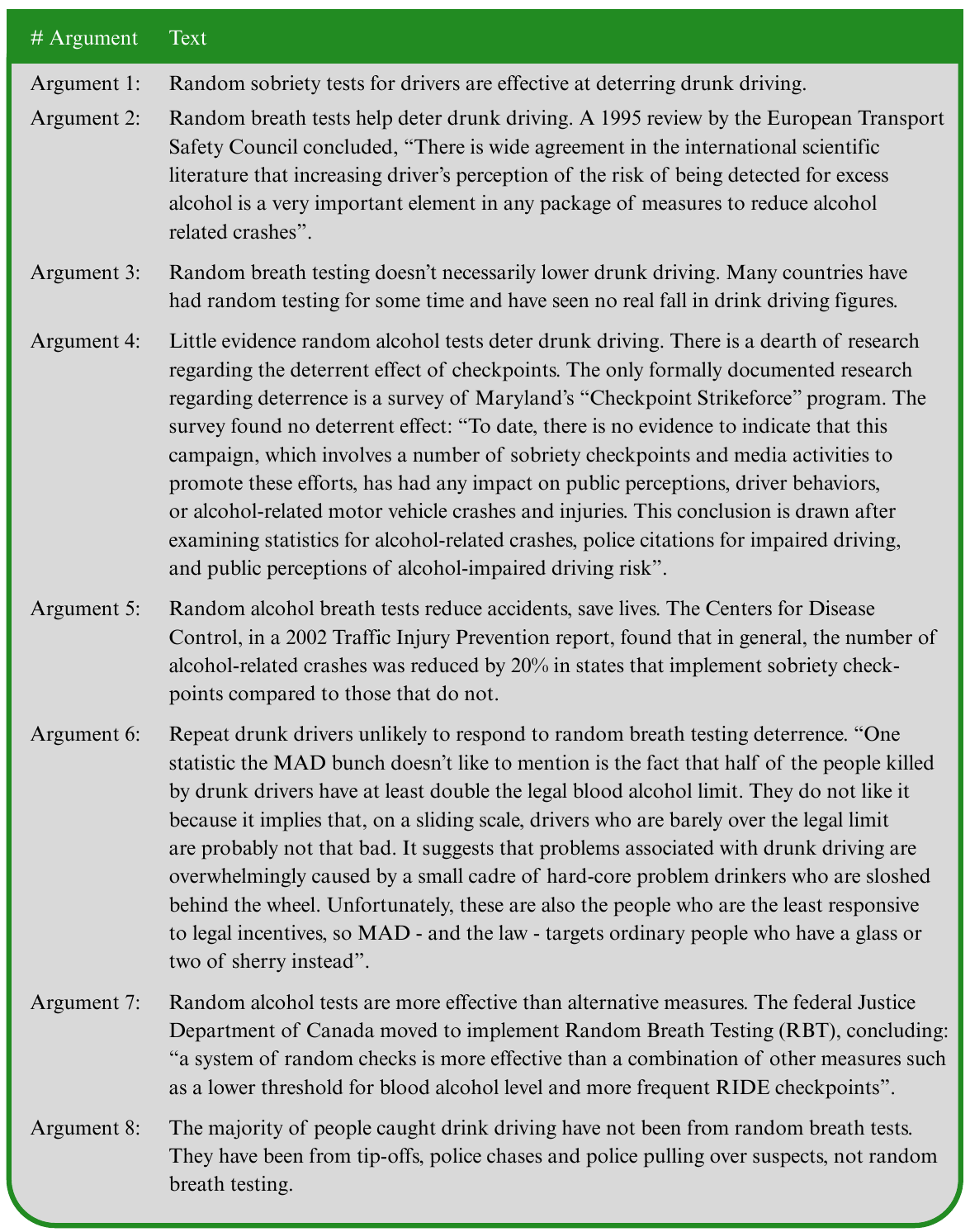}
        \caption{Chronologically ordered argument texts for the \textit{SobrietyTest} debate, derived from the XML representation in~\ref{fig:xml}. These correspond to the numbered nodes in the graph shown in Figure~\ref{fig:base_graph} and form the dialogue-style input presented to the LLM.}
        \label{fig:argtextpairs}
    \end{figure*}

    This flattening procedure preserves the narrative while masking the graph structure. The LLM sees only the surface form and must reason implicitly about the support/attack dynamics. Figure~\ref{fig:argtextpairs} shows the final dialogue-style representation used in our evaluation.

\section{Instruction Examples}
\label{sec:appendix-c}

We provide the main prompt templates used in our experiments here. Each corresponds to one of the instruction strategies used to evaluate LLMs' ability to reason over latent argument graphs. In all prompts, the placeholder \texttt{[Arguments]} refers to the dialogue-style input format illustrated in Appendix~\ref{app:graph-to-dialogue} (see Figure~\ref{fig:argtextpairs}).

    \subsection{Zero-Shot ``Vanilla'' Prompt}
    \label{app:vanilla}

\begin{figure}
        \centering
        \includegraphics[width=0.45\textwidth]{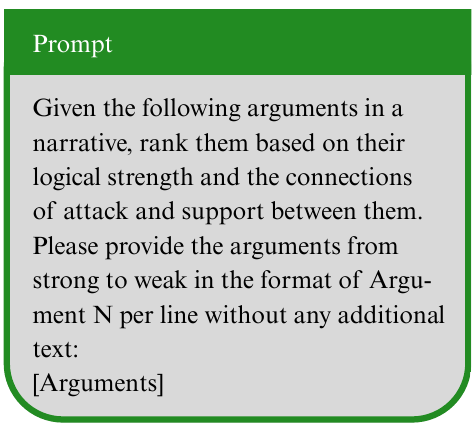}
        \caption{Zero-shot ``Vanilla'' instruction.}
        \label{fig:zer-shot-baseline-prompt}
    \end{figure}

    Figure~\ref{fig:zer-shot-baseline-prompt} shows the basic instruction prompt without any exemplars or reasoning steps.

    \subsection{One/Few-Shot ICL Prompt}
    \label{app:icl}

\begin{figure}
        \centering
        \includegraphics[width=0.45\textwidth]{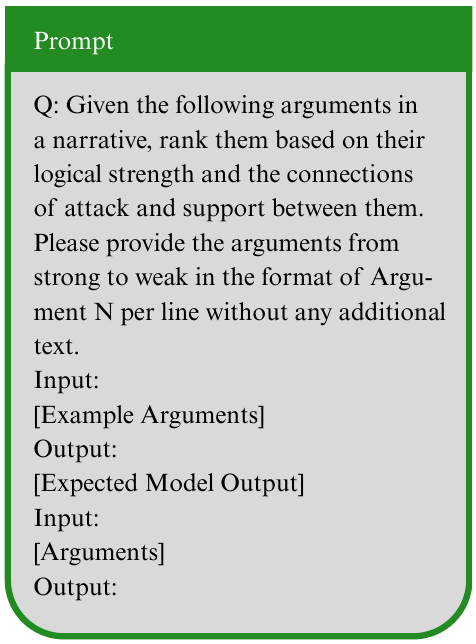}
        \caption{One-shot ICL instruction.}
        \label{fig:one-shot-icl-prompt}
    \end{figure}

    The ICL prompt shown in Figure~\ref{fig:one-shot-icl-prompt} includes one example; the few-shot variant uses three exemplars covering different graph types.

    \subsection{Zero-Shot CoT Prompt}
    \label{app:0-cot}

\begin{figure}
        \centering
        \includegraphics[width=0.45\textwidth]{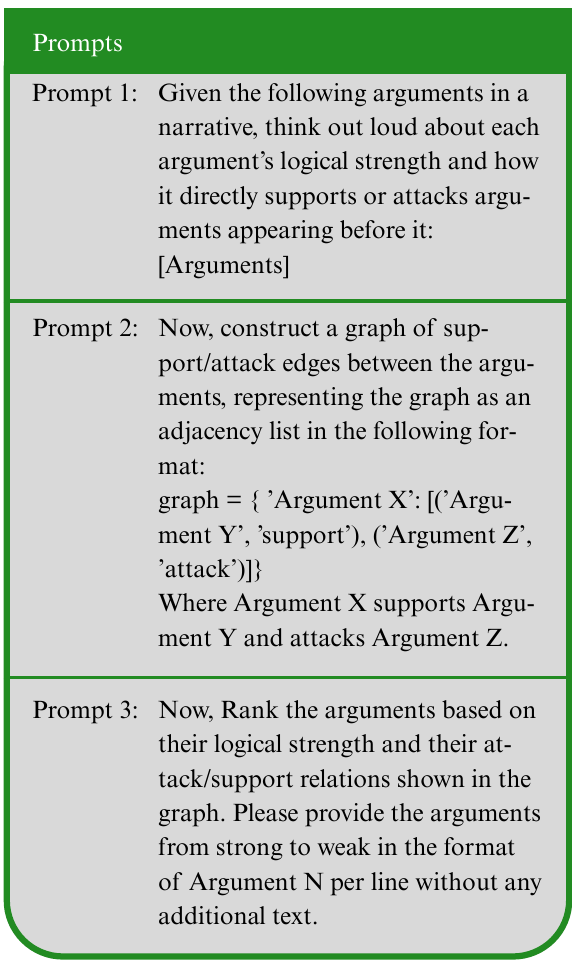}
        \caption{Zero-shot CoT instructions.}
        \label{fig:zero-cot-prompt}
    \end{figure}

    Figure~\ref{fig:zero-cot-prompt} presents the zero-shot CoT format, which prompts the model to reason step-by-step through the debate.

    \subsection{One/Few-Shot CoT Prompt}
    \label{app:few-cot}

\begin{figure}
    \centering
    \includegraphics[width=0.45\textwidth]{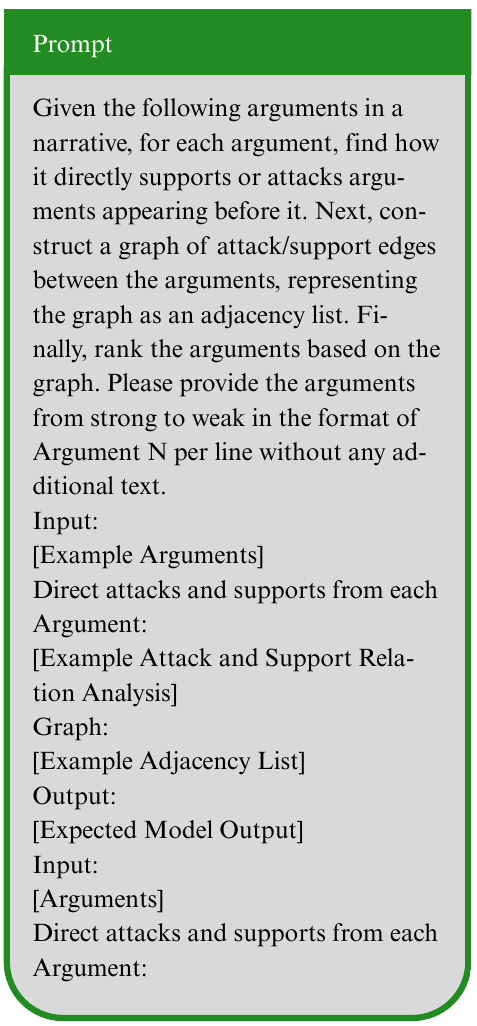}
    \caption{One-shot CoT instruction.}
    \label{fig:one-shot-cot-prompt}
\end{figure}

    As shown in Figure~\ref{fig:one-shot-cot-prompt}, the one-shot CoT prompt includes reasoning steps and edge inference. The few-shot variant extends this to three full exemplars.

\section{Debate Structure Reconstruction Detailed Results} \label{sec:appendix-d}

\begin{table}
    \centering
    \small
    \setlength{\tabcolsep}{6pt}
    \begin{tabular}{l l S[table-format=1.2] S[table-format=1.2] S[table-format=1.2]}
    \toprule
    \textbf{Model} & \textbf{Technique} & {\textbf{P}} & {\textbf{R}} & {\textbf{F1}} \\
    \midrule
    \multirow{3}{*}{gpt-4o}               & CoT Zero-Shot     & 0.36 & 0.75 & 0.47 \\
    & CoT One-Shot & 0.74 & 0.74 & \bfseries 0.74 \\
    & CoT Few-Shot & 0.73 & 0.73 & 0.73 \\
    \addlinespace
    \multirow{3}{*}{claude-3}             & CoT Zero-Shot     & 0.27 & 0.53 & 0.35 \\
                         & CoT One-Shot & 0.69 & 0.77 & 0.72 \\
                         & CoT Few-Shot & 0.73 & 0.72 & \bfseries 0.73 \\
    \addlinespace
    \multirow{3}{*}{cmd-r-plus}           & CoT Zero-Shot     & 0.06 & 0.22 & 0.09 \\
                         & CoT One-Shot & 0.09 & 0.08 & 0.09 \\
                         & CoT Few-Shot & 0.72 & 0.71 & \bfseries 0.72 \\   
    \addlinespace
    \multirow{3}{*}{llama-3}              & CoT Zero-Shot     & 0.27 & 0.42 & 0.33 \\
                         & CoT One-Shot & 0.62 & 0.64 & 0.62 \\
                         & CoT Few-Shot & 0.66 & 0.66 & \bfseries 0.66 \\
    \bottomrule
    \end{tabular}
    \caption{Adjacency-list recovery on \textit{DebatePedia}. Best F1 per model is bolded.}
    \label{tab:debatepedia-adj-appendix}
    \end{table}

\begin{table*}
    \centering
    \small
    \newcommand{\NA}{--}
    \begin{tabular}{l ccc ccc ccc}
      \toprule
      & \multicolumn{3}{c}{Act 1} & \multicolumn{3}{c}{Act 2} & \multicolumn{3}{c}{Act 3} \\
      \cmidrule(lr){2-4}\cmidrule(lr){5-7}\cmidrule(lr){8-10}
      Model & P & R & F1 & P & R & F1 & P & R & F1 \\
      \midrule
      \rowcolor{Gray}\multicolumn{10}{c}{CoT Zero-Shot} \\
      gpt-4o      & 0.67 & 0.42 & \textbf{0.52} & 0.66 & 0.66 & \textbf{0.66} & 0.47 & 0.90 & \textbf{0.62} \\
      claude-3    & 0.38 & 0.55 & 0.45 & 0.05 & 0.16 & 0.07 & 0.14 & 0.20 & 0.17 \\
      cmd-r-plus  & 0.25 & 0.34 & 0.29 & 0.09 & 0.22 & 0.12 & 0.20 & 0.50 & 0.29 \\
      llama-3     & 0.32 & 0.42 & 0.36 & 0.46 & 0.56 & 0.51 & 0.21 & 0.30 & 0.25 \\
      \rowcolor{Gray}\multicolumn{10}{c}{CoT One-Shot} \\
      gpt-4o      & 0.58 & 0.58 & \textbf{0.58} & 0.56 & 0.56 & \textbf{0.56} & 0.70 & 0.70 & 0.70 \\
      claude-3    & 0.50 & 0.53 & 0.51 & 0.42 & 0.56 & 0.48 & 0.71 & 1.00 & \textbf{0.83} \\
      cmd-r-plus  & \NA  & \NA  & \NA  & \NA  & \NA  & \NA  & \NA  & \NA  & \NA  \\
      llama-3     & 0.58 & 0.58 & \textbf{0.58} & 0.42 & 0.56 & 0.48 & 0.60 & 0.60 & 0.60 \\
      \rowcolor{Gray}\multicolumn{10}{c}{CoT Few-Shot} \\
      gpt-4o      & 0.39 & 0.39 & 0.39 & 0.63 & 0.63 & \textbf{0.63} & 0.90 & 0.90 & \textbf{0.90} \\
      claude-3    & 0.44 & 0.58 & 0.50 & 0.52 & 0.50 & 0.51 & 0.50 & 0.50 & 0.50 \\
      cmd-r-plus  & 0.24 & 0.58 & 0.34 & 0.37 & 0.63 & 0.47 & 0.50 & 0.70 & 0.58 \\
      llama-3     & 0.55 & 0.55 & \textbf{0.55} & 0.56 & 0.56 & 0.56 & 0.80 & 0.80 & 0.80 \\
      \bottomrule
    \end{tabular}
    \caption{Adjacency-list recovery on the three acts of \textit{12AngryMen}. Only the best F1 per act/setting is bolded. Blanks for \texttt{cmd-r-plus} in CoT one-shot are due to no valid adjacency list.}
    \label{tab:12am-adj-appendix}
\end{table*}

Tables~\ref{tab:debatepedia-adj-appendix} and~\ref{tab:12am-adj-appendix} 
show the detailed breakdown of adjacency-list recovery results for the 12AngryMen and DebatePedia datasets.

\section{DeepSeek R1 vs.\ General-Purpose Models (Plots)}\label{sec:appendix-e}
\begin{figure*}
    \centering
    \begin{subfigure}{0.47\textwidth}
        \includegraphics[width=\linewidth]{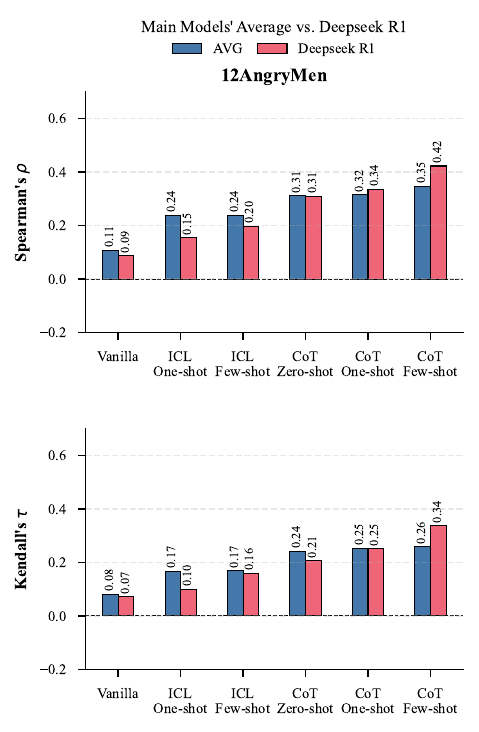}   
        \caption{Comparison on 12AngryMen dataset}
        \label{fig:deepseek_12angrymen}
    \end{subfigure}
    \hspace{2mm}
    \begin{subfigure}{0.47\textwidth}
        \includegraphics[width=\linewidth]{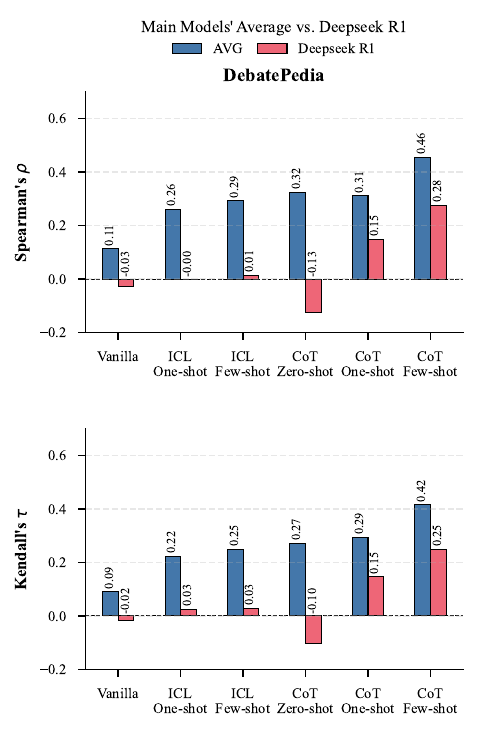}
        \caption{Comparison on DebatePedia dataset}
        \label{fig:deepseek_debatepedia}
    \end{subfigure}
    \caption{Argument ranking performance of DeepSeek R1 vs.\ the average of four general-purpose models across instruction settings; bars show Spearman's $\rho$ and Kendall's $\tau$. \textbf{(a)} \textit{12AngryMen}: R1 is comparable to the average of general-purpose models and benefits from one and few-shot CoT. \textbf{(b)} \textit{DebatePedia}: R1 underperforms all general-purpose models; CoT exemplars improve format adherence and recall but do not close the gap.}
    \label{fig:deepseek_comparison}
\end{figure*}
Figure~\ref{fig:deepseek_comparison} compares \emph{DeepSeek R1} to the \emph{average} of our four general-purpose models across instruction setups on \textit{12AngryMen} and \textit{DebatePedia}, reporting Spearman~$\rho$ (top) and Kendall~$\tau$ (bottom).

\end{document}